\documentclass[english]{article}
\usepackage[utf8]{inputenc}
\usepackage[T1]{fontenc}
\usepackage{babel}
\usepackage{amsmath}
\usepackage[superscript,nomove]{cite}
\usepackage{graphicx}
\graphicspath{ {./} }
\usepackage{fancyhdr}
\usepackage{changepage}
\usepackage{booktabs}
\usepackage[hyphenbreaks]{breakurl}
\usepackage[hyphens]{url}
\usepackage{doi}
\begin{document}

\title{Question-Driven Summarization of Answers to Consumer Health Questions}
\date{}
\author{Max Savery\textsuperscript{1}, Asma Ben Abacha\textsuperscript{1}, Soumya Gayen\textsuperscript{1}, Dina Demner-Fushman\textsuperscript{1{*}}}

\maketitle
\thispagestyle{fancy}

1.  Lister Hill National Center for Biomedical Communications, U.S. National Library of Medicine, National Institutes of Health, Bethesda, MD. {*}corresponding author(s):
Dina Demner-Fushman (ddemner@mail.nih.gov)
\begin{abstract}

Automatic summarization of natural language is a widely studied area in computer science, one that is broadly applicable to anyone who routinely needs to understand large quantities of information. For example, in the medical domain, recent developments in deep learning approaches to automatic summarization have the potential to make health information more easily accessible to patients and consumers. However, to evaluate the quality of automatically generated summaries of health information, gold-standard, human generated summaries are required. Using answers provided by the National Library of Medicine’s consumer health question answering system, we present the MEDIQA-Answer Summarization dataset, the first summarization collection containing question-driven summaries of answers to consumer health questions. This dataset can be used to evaluate single or multi-document summaries generated by algorithms using extractive or abstractive approaches. In order to benchmark the dataset, we include results of baseline and state-of-the-art deep learning summarization models, demonstrating that this dataset can be used to effectively evaluate question-driven machine-generated summaries and promote further machine learning research in medical question answering.

\end{abstract}

\section*{Background \& Summary}
Searching online for health information can be difficult for even the most savvy of users. A conventional search engine will return a set of web pages in response to a query, but the consumer of this health information is not always able to judge the correctness and relevance of the content\cite{NLMCHinterview}. To make it easier to find and understand online health information, the National Library of Medicine (NLM) has developed the Consumer Health information Question Answering (CHiQA)\cite{Demner-Fushman2019} online system, shown in Figure 1. CHiQA (available at https://chiqa.nlm.nih.gov) indexes only pages from sites hosted by reliable organizations, such as MedlinePlus, Mayo Clinic, and other resources maintained by the National Institutes of Health. In response to consumers' health questions, CHiQA provides passages from these pages as answers, using a variety of information retrieval and machine learning techniques. 

While retrieving reliable information is important, it is also essential that the consumer be able to easily and quickly understand this information. To this end, recent developments in automatic text summarization, a field at the intersection of machine learning and natural language processing, have shown the potential to aid consumers in understanding health information\cite{summqs}. However, to develop more advanced summarization tools for medical question answering systems such as CHiQA, there needs to exist human-curated datasets that can be used for evaluating automatically generated summaries. Additionally, we are particularly interested in summaries that are \textit{question-driven}, i.e., summaries that contain information relevant to helping the consumer answer their question. Therefore, in order to promote summarization research in the medical domain, we propose a new gold-standard dataset, MEDIQA-Answer Summarization (MEDIQA-AnS)\cite{savery-datacit} for the evaluation of question-driven summaries of health information.

\begin{figure}[hbtp]
\centering
\includegraphics[scale=.5]{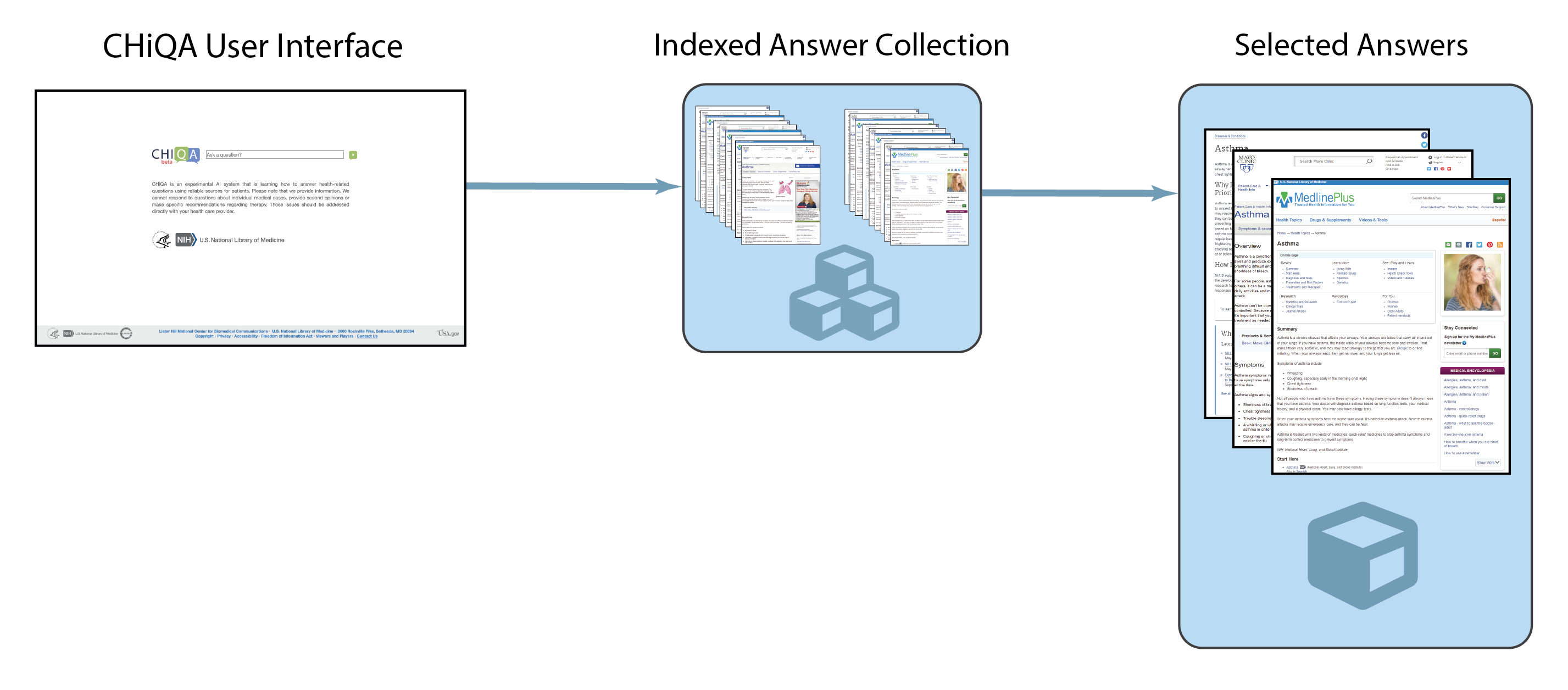}
\caption{CHiQA interface and pipeline.}
\end{figure}

Popular summarization datasets include the CNN-Dailymail dataset\cite{hermann2015}, which uses headlines as summaries of news articles, and the PubMed dataset\cite{cohan2018}, which uses abstracts as summaries of scientific articles. These can be used for training and testing summarization systems on short open domain text and long scientific text, respectively. Additionally, datasets such as Multi-news\cite{fabbri2019}, BioASQ\cite{bioasq2015}, and SciSumm \cite{scisumm} can be used for further types of single and multi-document summarization evaluations, but not for evaluating question-driven, multi-document summaries generated by medical summarization systems.

Recently, MEDIQA 2019 shared task\cite{MEDIQA2019} introduced the MEDIQA-QA dataset for answer-ranking, encouraging research in medical question answering systems. This dataset consists of consumer health questions and passages selected from reliable online sources, making it uniquely suited for the purpose of this paper: For each question, the passages can be considered a question-driven summary of the original web page. We therefore used the text of the web pages and the extracted passages as the base data source for MEDIQA-AnS. 

To extend MEDIQA-QA for answer summarization, we manually generated single and multi-document summaries of the passages. We created two versions of each summary: An extractive version, where the summary consisted of content copied-and-pasted from the source text, and an abstractive version, where the summary was written from scratch, using the source text as reference. The unique structure of this dataset allows for the evaluation of many different types of summarization systems, including those focused on single document or multi-document summarization, long or short document, and extractive or abstractive approaches. In addition to releasing MEDIQA-AnS, we include results from baseline and state-of-the-art approaches to medical answer summarization, focusing on the single document summarization aspect of the task, in order to benchmark the dataset for future researchers. 

\begin{table}[h]
\centering
\caption{Frequency of reliable websites included in MEDIQA-AnS.}
\begin{tabular}{lc}
\toprule
Website                   & Frequency \\
\midrule
    medlineplus.gov           & 190 \\
    mayoclinic.org        & 151 \\
    nlm.nih.gov           & 44  \\
    rarediseases.info.nih.gov & 39  \\
    ghr.nlm.nih.gov           & 31  \\
    nhlbi.nih.gov         & 22  \\
    niddk.nih.gov         & 21  \\
    ninds.nih.gov         & 16  \\
    womenshealth.gov      & 12  \\
    nihseniorhealth.gov       & 8   \\
    nichd.nih.gov         & 7   \\
    niams.nih.gov         & 6   \\
    cancer.gov            & 3   \\
    nia.nih.gov           & 1   \\
    nei.nih.gov               & 1  \\
\hline
\bottomrule
\end{tabular}
\end{table}

\section*{Methods}
To create the original MEDIQA-QA dataset, consumer health questions were submitted to CHiQA, and for each question a set of answers was returned. From the full text of the web pages containing the answers, passages containing related information were selected and the relevance of each passage was manually rated. Importantly, it is these passages that we summarized for the MEDIQA-AnS collection.

To create the summaries, we first filtered passages from MEDIQA-QA which had been assessed as incomplete or excellent. Table 1 shows the frequency of source websites that this subset of answers was selected from. Then, for each question and corresponding set of answers, we generated the following types of summaries:

\begin{itemize}
\item Extractive summary of each answer 
\item Abstractive summary of each answer 
\item Multi-document extractive summary considering the information presented in all of the answers
\item Multi-document abstractive summary
\end{itemize}

\begin{table}[h]
\centering
\caption{ROUGE-2, ROUGE-L and BLEU inter-annotator agreement for each summary type.}
\begin{tabular}{lccc}
\toprule
    Summarization type &   ROUGE-2 &   ROUGE-L &      BLEU \\
\midrule
  Multi-document, abstractive &  0.19 &  0.32 &  0.17 \\
   Multi-document, extractive &  0.56 &  0.57 &  0.49 \\
 Single document, abstractive &  0.28 &  0.42 &  0.19 \\
  Single document Extractive &  0.82 &  0.83 &  0.74 \\
\hline
\bottomrule
\end{tabular}
\end{table}

\begin{table}[h]
\centering
\caption{ROUGE-2 and BLEU calculated using abstractive summaries as the reference summary and the extractive summaries as the candidate summary.}
\begin{tabular}{lcc}
\toprule
Summary type                              & ROUGE-2 & BLEU \\
\midrule
Single document, abstractive v. extractive & 0.64      & 0.41   \\
Multi-document abstractive, v. extractive  & 0.62      & 0.42 \\
\hline 
\bottomrule
\end{tabular}
\end{table}

The summaries of the answers were generated by two subject matter experts, using a summarization interface we created to allow the annotators to input separate summaries of each type. The extractive summaries were generated by selecting chunks of text from the answers. Though the source text was sometimes punctuated correctly, it also included lists, headings, and other types of formatting. We selected text regardless of formatting, considering which chunks contained the relevant information.

For the abstractive summaries, the source text or extractive summary was rewritten into easier to understand, condensed sentences. Writing the abstractive summaries involved either rewording chunks, reorganizing and reformatting sentences, or potentially using extracted text that was already clear and informative. Since the answers were selected from reliable online medical resources, there were many cases in which the extractive summary was already well-worded and clear. Finally, once the extractive and abstractive summaries were written, multi-document summaries were created using all of the answers.

ROUGE\cite{rouge} and BLEU\cite{bleu} were used to calculate inter-annotator agreement, as shown in Table 2. The inter-annotator measures show that the annotators more frequently use the same n-grams when creating extractive summaries. This is to be expected, as it is less likely that two individuals will use the exact same combinations of words when generating novel, abstractive text.

Additionally, we wanted to measure the similarity between the abstractive and extractive summaries. We used ROUGE-2 and BLEU as a similarity measure, using the abstractive summaries as the reference summary and the extracted summaries as the candidates. The scores shown in Table 3 indicate that the extractive summaries do contain many of the same n-grams as the abstractive summaries. However, it is important to note here that even if a pair of summaries receive a low ROUGE or BLEU score, neither are necessarily incorrect, poorly written, or lacking information. To demonstrate this point, Figure 2 shows a consumer health question, two summaries written by different annotators, and the ROUGE-2 score between the summaries. While the ROUGE-2 score is quite low, both summaries contain information relevant to the question, the main difference between the two summaries being that one focuses on genetics, the other on family history. This highlights the inability of ROUGE, and relatedly BLEU, to capture the nuanced differences between two otherwise informative summaries. Further discussion of these metrics is included in the technical validation section.

\begin{figure}[h]
\centering
\includegraphics[scale=.85]{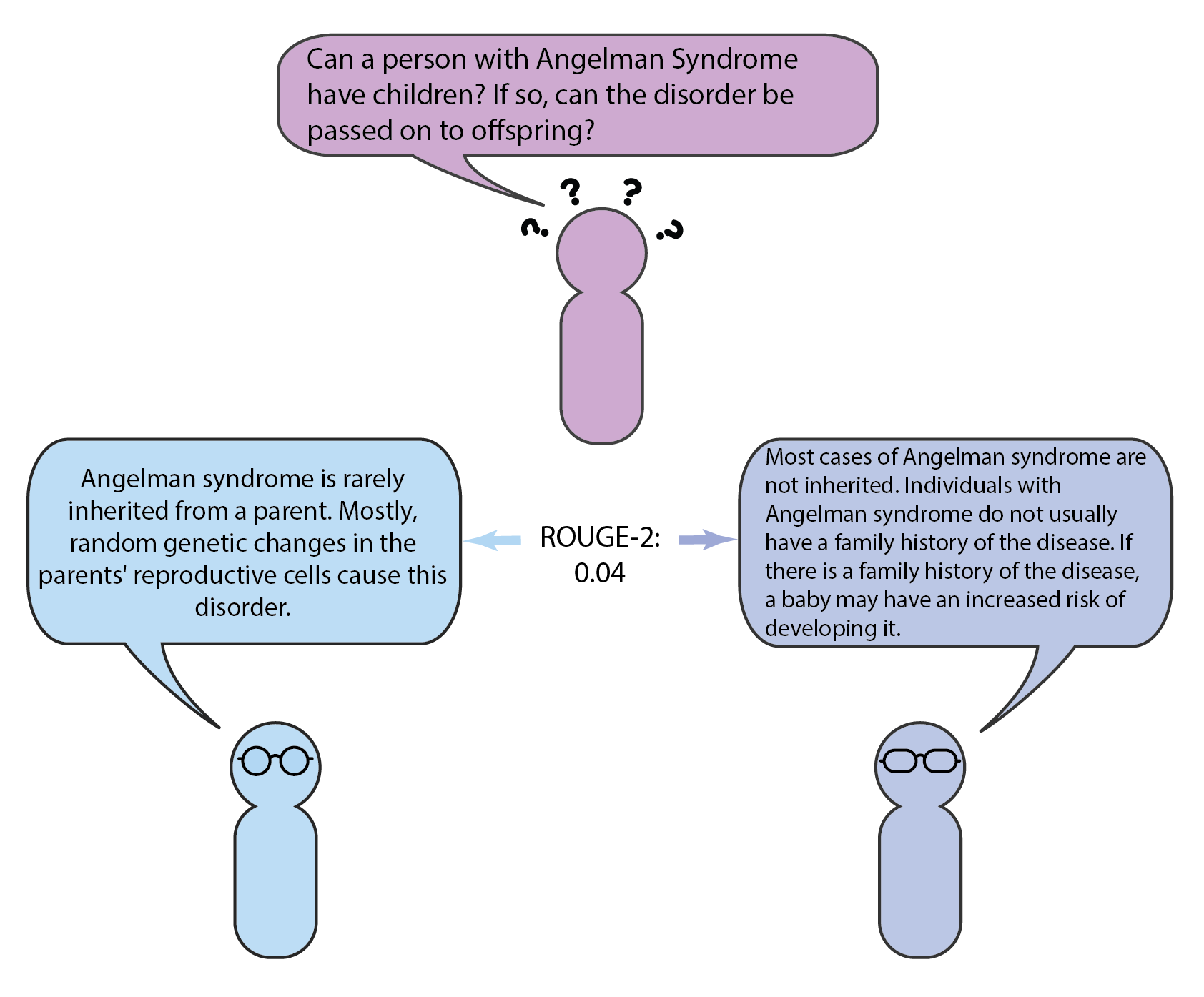}
\caption{Comparison of summaries of an answer to a consumer health question. While both summaries are relevant, they receive a low ROUGE-2 score.}
\end{figure}

\subsection*{Code availability}
The code to reconstruct the data for this paper and to reproduce the results shown here can be found at \burl{https://www.github.com/saverymax/qdriven-chiqa-summarization}.

\vspace{10pt}
\begin{figure}[h]
\centering
\includegraphics[scale=.5]{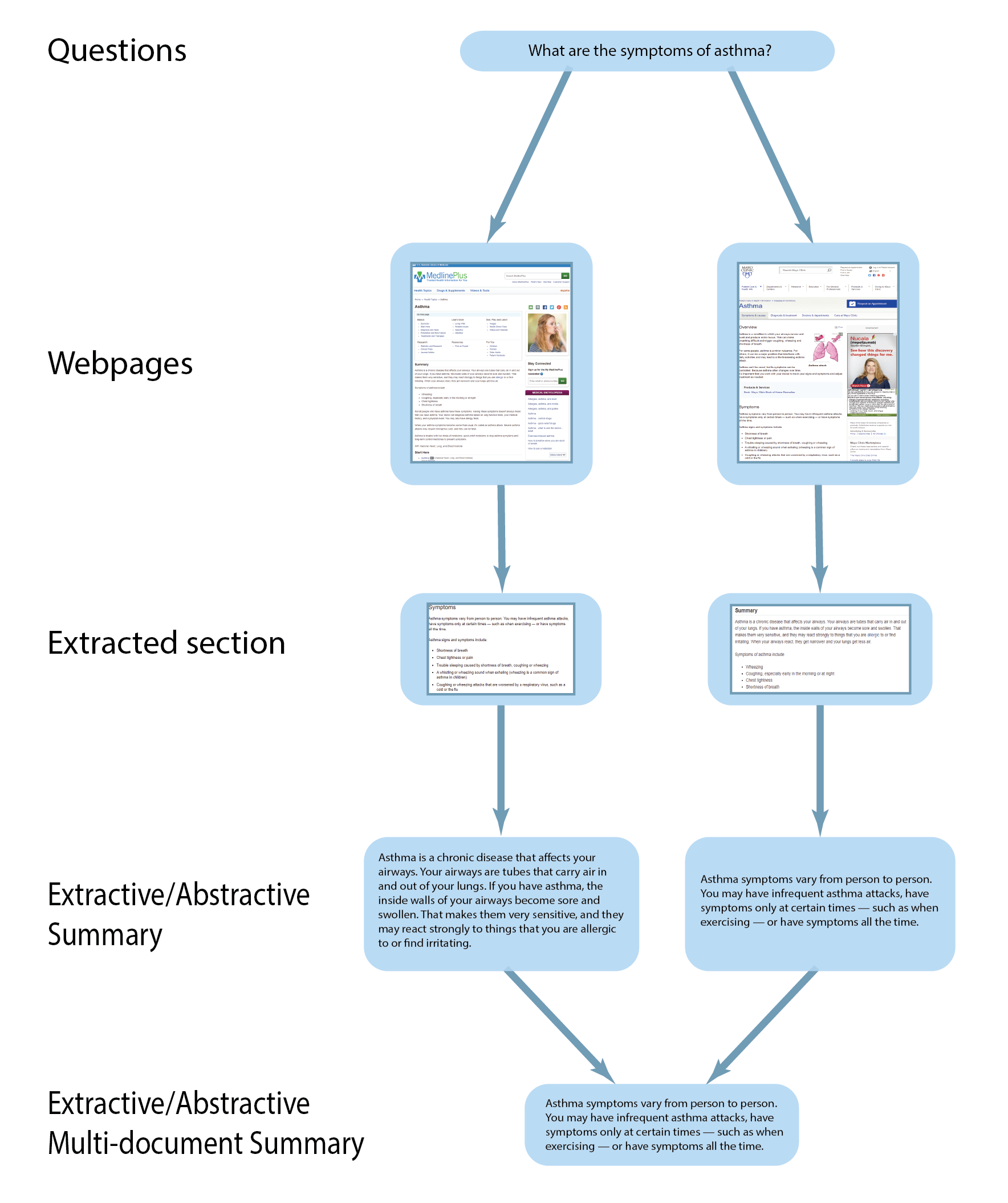}
\caption{Example of a data record in MEDIQA-AnS.}
\end{figure}

\section*{Data Records}
We have archived nine data records with Open Science Framework (OSF), available at \doi{10.17605/OSF.IO/FYG46}\cite{savery-datacit}. The primary dataset contains 156 questions, the text of the web pages returned by CHiQA, the passages selected from the web pages, and the abstractive and extractive, multi and single document summaries of the passages. The additional eight datasets are subsets of this dataset, divided into potential experimental use cases. For example, we have included a split containing questions and corresponding web pages, and the multi-document summaries of these pages. This allows users to directly evaluate a system on multi-document summarization without having to perform additional data processing on the whole dataset.

\begin{table}[h]
\centering
\caption{Average tokens and sentences per document type.}
%\begin{adjustwidth}{-3cm}{}
%\begin{tabular}{ lc|c@{}c|c@{}c }% <-- aded @{}
\begin{tabular}{lccccccc}
\toprule
&& \multicolumn{2}{c}{Tokens} & \multicolumn{2}{c}{Sentences} \\
\hline
%&& Tokens && Sentences & \\

Data type                                & Count & Average  & SD  & Average  &    SD  \\
\midrule
Questions                    & 156    & 25             & 31          & 2                        & 2              \\
Unique articles    & 348    & 1675           & 1798        & 95                      & 104            \\
Passages                     & 552    & 631            & 869         & 35                       & 48             \\
\hline
Summaries &&&&& \\
\hline
Multi-document abstractive  & 156    & 141            & 119         & 7                        & 6              \\
Multi-document extractive   & 156    & 220            & 183         & 12                       & 12             \\
Single-document abstractive & 552    & 83             & 78          & 4                        & 4              \\
Single-document extractive  & 552    & 133            & 127         & 7                        & 8             \\
\hline
\bottomrule
\end{tabular}
%\end{adjustwidth}
\end{table}

Each dataset is saved in JSON format, where each key is a question id and each value contains a nested JSON object with the question, web pages, passages, summaries, rating of web pages from MEDIQA-QA, and the URL for each web page. A summary of the structure of the data for a single question is shown in Figure 3, and statistics regarding the questions, articles, and summaries are shown in Table 4. More detailed descriptions regarding the potential use of each dataset and their respective key and value pairs can be found in the README file in the OSF archive.

\section*{Technical Validation}
To benchmark the MEDIQA-AnS dataset and to demonstrate how it can be used to evaluate automatically generated summaries, we conducted a series of experiments using a variety of summarization approaches. Three baseline and three deep learning algorithms were implemented and are listed below: 
\begin{description}

\item \textit{Lead-3}: The Lead-3 baseline takes the first three sentences of an article as a summary. This has been shown in previous work\cite{Pgen,Nallapati-summa} to be an effective baseline for summarization and is the current approach implemented in CHiQA. 

\item \textit{k random sentences}: Similarly to the Lead-3 baseline, we select $k=3$ random sentences from each article.

\item \textit{k-best ROUGE}: We select $k=3$ sentences with the highest ROUGE-L score relative to the question.

\item \textit{Bidirectional Long Short-Term Memory (BiLSTM) network}: A BiLSTM\cite{lstm} was trained to select the most relevant sentences in an article, similar to other extractive LSTM models\cite{chen2018fast,Liu2019HierarchicalTF}.

\item  \textit{Pointer-Generator network}: The Pointer-Generator network\cite{Pgen} is a hybrid sequence-to-sequence attentional model, with the capability to create summaries by copying text from the source document while also generating novel text.

\item \textit{Bidirectional Autoregressive Transformer (BART)}: BART\cite{lewis2019bart} is a recently published generative transformer-based model combining the bidrectional encoder of BERT\cite{devlin2018} with the autoregressive decoder of GPT-2\cite{radford2019}. 
\end{description}

All models were trained using the questions, abstracts, and snippets available in the BioASQ data. The BioASQ data can be easily adapted for training summarization models, as this is one of the tasks included in the BioASQ challenge. For validation of loss during training, we used the medical question and answer dataset MedInfo\cite{BenAbacha:MEDINFO19}. This dataset consists of answers selected from reliable online health information, in response to consumer health questions about medications. It is therefore similar in structure and content to the MEDIQA-QA data, and can be used to approximate the single document, extractive summarization task provided in the MEDIQA-AnS collection.

We use these methods to automatically summarize the full text of the web pages in MEDIQA-AnS. Table 5 and 6 show the comparison between the automatically generated summaries and the manually generated summaries. We include results for only single document summarization; however, the same experiments could be run in a multi-document setting.

\begin{table}[h]
\centering
\caption{Automatically generated summaries compared to extractive summaries.}
\begin{tabular}{lcccc}
\toprule
    Experiment &   ROUGE-1 &   ROUGE-2 &   ROUGE-L &      BLEU \\
\midrule
    Lead-3  &  0.23 &  0.11 &  0.08 &  0.04 \\
    3-random &  0.20 &  0.08 &  0.06 &  0.04 \\
    3-best ROUGE &  0.16 &  0.08 &  0.06 &  0.00 \\
    BiLSTM &  0.22 &  0.10 &  0.08 &  0.03 \\
    Pointer-Generator &  0.21 &  0.09 &  0.07 &  0.03 \\
    BART &  0.29 &  0.15 &  0.12 &  0.09 \\
\hline
\bottomrule
\end{tabular}

\end{table}

\begin{table}[h]
\centering
\caption{Automatically generated summaries compared to abstractive summaries.}

\begin{tabular}{lcccc}
\toprule
    Experiment &   ROUGE-1 &   ROUGE-2 &   ROUGE-L &      BLEU \\
\midrule
    Lead-3 &  0.25 &  0.10 &  0.07 &  0.06 \\
    3-random &  0.22 &  0.07 &  0.04 &  0.05 \\
    3-best ROUGE &  0.17 &  0.07 &  0.04 &  0.02 \\
    BiLSTM &  0.24 &  0.09 &  0.06 &  0.06 \\
    Pointer-Generator &  0.24 &  0.08 &  0.06 &  0.05 \\
    BART &  0.32 &  0.12 &  0.08 &  0.09 \\
\hline
\bottomrule
\end{tabular}
\end{table}

The Lead-3 baseline scores well compared to the other methods, as expected. Relevant to machine learning research, BART outperforms the Pointer-Generator (\textit{p}<0.0039, across all experiments, Wilcoxon signed-rank test), which is consistent with previous work\cite{lewis2019bart}. Additionally, it is interesting to note that the ROUGE-1 and BLEU scores for each model tend to increase in the abstractive evaluation, and the ROUGE-2 and ROUGE-L scores tend to increase in the extractive evaluation. The increase in ROUGE-1 is potentially due to the fact that, as indicated in Table 4, the abstractive summaries are shorter, which makes it easier for an automatically generated summary to contain a larger percentage of unigrams present in the reference summary. The increase in ROUGE-2 and ROUGE-L in the extractive evaluation is likely because it is easier for a system to get a sequence of tokens correct in an extractive setting. These observations may be useful to consider while using MEDIQA-AnS, since the type of summary being used for evaluation will affect the observed performance of a system. 

As mentioned previously, when using ROUGE and BLEU scores to make a judgement about the quality of generated summaries it is important to take account of the way they are calculated: The metrics measure the number of n-grams occurring in a candidate summary when compared to a reference summary, where ROUGE is oriented for recall and BLEU is oriented for precision. A high-quality candidate summary may contain some of the n-grams present in the reference summary, but there is no requisite that it must contain the exact same wording, as illustrated in Figure 2. When performing evaluations with MEDIQA-AnS, the bias of these metrics should be carefully considered. It may be useful for researchers to conduct a manual error analysis in addition to calculating the automatic metrics, as a system might not be producing a high-quality summary just because it is achieving a high ROUGE or BLEU score. 

\begin{table}[h]
\centering
\caption{BART using question-driven approach. Shows summaries generated with and without access to the question, compared to extractive and abstractive summaries. Across all experiments, BART scores higher when the question is used as input}

\begin{tabular}{lcccc}
\toprule
                                        Experiment &   ROUGE-1 &   ROUGE-2 &   ROUGE-L &    BLEU \\
\midrule
\newline
Pages \\
%\newline
              BART+Q, Abstractive &  0.32 &  0.12 &  0.08 &  0.09 \\
           BART, Abstractive &  0.26 &  0.09 &  0.05 &  0.07 \\
               BART+Q, Extractive &  0.29 &  0.15 &  0.12 &  0.09 \\
            BART, Extractive &  0.24 &  0.10 &  0.07 &  0.05 \\
\midrule
\newline           
Passages \\

              BART+Q, Abstractive &  0.46 &  0.29 &  0.24 &  0.19 \\
            BART, Abstractive &  0.43 &  0.27 &  0.21 &  0.17 \\
                BART+Q, Extractive &  0.46 &  0.37 &  0.35 &  0.18 \\
           BART, Extractive &  0.43 &  0.35 &  0.33 &  0.14 \\
\hline
\bottomrule
\end{tabular}
%\end{adjustwidth}
\end{table}

In order to determine if the dataset could be used for the evaluation of question-driven summarization, we trained BART with and without access to the consumer health questions. For training and testing in these experiments, we concatenated the question to the beginning of the article. This approach is similar to other deep learning text generation work, where including unique text at the beginning of the documents fed to a model can give greater control over the content of the output. The models learn to associate these additional strings with certain information during training. For example, users can provide the CTRL model\cite{ctrl} with control codes to specify the topic of generated text.

After including the question with the text during training and evaluation, we observed a significant difference between the two types of summaries. Table 7 shows that including the question with the input documents significantly improved BART's performance across all summarization tasks: passages and pages, extractive and abstractive (\textit{p}<0.0064 across all experiments, Wilcoxon signed-rank test). This indicates that being able to compare plain summaries with question-driven summaries is important to developing medical question answering systems, and since there are no existing datasets for this type of evaluation, MEDIQA-AnS can play a valuable role for further research in this area.

\section*{Usage Notes}
We have provided instructions in the README file in the Open Science Framework describing how to process the MEDIQA-AnS dataset. The code for processing the data and evaluating the summarization systems shown here can be found in the scripts located at the GitHub repository provided above.

\section*{Acknowledgements}
This research was supported by the Intramural Research Program of the National Institutes of Health, National
Library of Medicine, and Lister Hill National Center for Biomedical Communications.

\section*{Author contributions}
M.S implemented data processing code and pipelines, conducted baseline and machine learning experiments, contributed to generating the manual summaries, and wrote and edited the manuscript. A.B-A authored the MEDIQA data used as the backbone for the collection presented here, as well as the MedInfo data used for training, provided guidance on their use, developed the summarization interface, edited the manuscript, and provided feedback for collection creation. S.G managed the interface for generating the summaries, and provided data processing support. D.D-F conceived of the project, contributed to generating the manual summaries, edited the manuscript, and otherwise provided feedback on all aspects of the study. 

\section*{Competing interests}
The authors declare no competing financial interests.

%\bibliography{asumm}{}

\begin{thebibliography}{10}
\expandafter\ifx\csname url\endcsname\relax
  \def\url#1{\texttt{#1}}\fi
\expandafter\ifx\csname urlprefix\endcsname\relax\def\urlprefix{URL }\fi
\providecommand{\bibinfo}[2]{#2}
\providecommand{\eprint}[2][]{\url{#2}}

\bibitem{NLMCHinterview}
\bibinfo{author}{National Network of Libraries of Medicine}
\newblock \bibinfo{title}{The consumer health reference interview and ethical
  issues}. https://nnlm.gov/initiatives/topics/ethics.

\bibitem{Demner-Fushman2019}
\bibinfo{author}{Demner-Fushman, D.}, \bibinfo{author}{Mrabet, Y.} \&
  \bibinfo{author}{Ben~Abacha, A.}
\newblock \bibinfo{title}{{Consumer health information and question answering:
  helping consumers find answers to their health-related information needs}}.
\newblock \emph{\bibinfo{journal}{Journal of the American Medical Informatics
  Association}} \textbf{\bibinfo{volume}{27}}, \bibinfo{pages}{194--201}
  (\bibinfo{year}{2019}).

\bibitem{summqs}
\bibinfo{author}{Demner-Fushman, D.} \&
  \bibinfo{author}{Ben~Abacha, A.}
\newblock \bibinfo{title}{{On the Summarization of Consumer Health Questions}}.
  \newblock In \emph{\bibinfo{booktitle}{Proceedings of the 57th Annual Meeting of the Association for Computational Linguistics}},
  \bibinfo{pages}{2228--2234} (\bibinfo{publisher}{Association for Computation Linguistics}, 
  \bibinfo{address}{Florence, Italy}
  \bibinfo{year}{2019}).

\bibitem{savery-datacit}
\bibinfo{author}{Savery, M.}, \bibinfo{author}{{Ben Abacha}, A.},
  \bibinfo{author}{Gayen, S.} \& \bibinfo{author}{Demner{-}Fushman, D.}
\newblock \bibinfo{title}{Question-driven summarization of answers to consumer
  health questions} \url{https://doi.org/10.17605/OSF.IO/FYG46} (\bibinfo{year}{2020}).

\bibitem{Pgen}
\bibinfo{author}{See, A.}, \bibinfo{author}{Liu, P.~J.} \&
  \bibinfo{author}{Manning, C.~D.}
  \newblock \bibinfo{title}{Get to the point: summarization with
  pointer-generator networks}.
  \newblock In \emph{\bibinfo{booktitle}{Proceedings of the 55th Annual Meeting of the Association for Computational Linguistics}},
  \bibinfo{pages}{1073–1083} (\bibinfo{publisher}{Association for Computation Linguistics}, 
  \bibinfo{address}{Vancouver, Canada}
  \bibinfo{year}{2017}).

\bibitem{Nallapati-summa}
\bibinfo{author}{Nallapati, R.}, \bibinfo{author}{Zhai, F.} \&
  \bibinfo{author}{Zhou, B.}
\newblock \bibinfo{title}{Summarunner: a recurrent neural network based
  sequence model for extractive summarization of documents}.
  \newblock In \emph{\bibinfo{booktitle}{AAAI Conference on Artificial Intelligence}}
  (\bibinfo{year}{2017}).

\bibitem{hermann2015}
\bibinfo{author}{Hermann, K.~M.} \emph{et~al.}
\newblock \bibinfo{title}{Teaching machines to read and comprehend}.
\newblock In \emph{\bibinfo{booktitle}{Advances in neural information
  processing systems}}, \bibinfo{pages}{1693--1701} (\bibinfo{year}{2015}).

\bibitem{cohan2018}
\bibinfo{author}{Cohan, A.} \emph{et~al.}
\newblock \bibinfo{title}{A discourse-aware attention model for abstractive
  summarization of long documents}.
  \newblock In \emph{\bibinfo{booktitle}{Proceedings of the 2018 Conference of the North {A}merican Chapter of the Association for Computational Linguistics}},
  \bibinfo{pages}{615--621} (\bibinfo{publisher}{Association for Computation Linguistics}, 
  \bibinfo{address}{New Orleans, Louisiana}
  \bibinfo{year}{2018}).

\bibitem{fabbri2019}
\bibinfo{author}{Fabbri, A.~R.}, \bibinfo{author}{Li, I.},
  \bibinfo{author}{She, T.}, \bibinfo{author}{Li, S.} \&
  \bibinfo{author}{Radev, D.~R.}
\newblock \bibinfo{title}{Multi-news: a large-scale multi-document
  summarization dataset and abstractive hierarchical model}.
  \newblock In \emph{\bibinfo{booktitle}{Proceedings of the 57th Annual Meeting of the Association for Computational Linguistics}},
  \bibinfo{pages}{1074--1084} (\bibinfo{publisher}{Association for Computation Linguistics}, 
  \bibinfo{address}{Florence, Italy}
  \bibinfo{year}{2019}).

\bibitem{bioasq2015}
\bibinfo{author}{Tsatsaronis, G.} \emph{et~al.}
\newblock \bibinfo{title}{An overview of the bioasq large-scale biomedical
  semantic indexing and question answering competition}.
\newblock \emph{\bibinfo{journal}{BMC Bioinformatics}}
  \textbf{\bibinfo{volume}{16}}, \bibinfo{pages}{138} (\bibinfo{year}{2015}).

\bibitem{scisumm}
\bibinfo{author}{Yasunaga, M.} \emph{et~al.}
\newblock \bibinfo{title}{{ScisummNet}: a large annotated corpus and
  content-impact models for scientific paper summarization with citation
  networks}.
\newblock In \emph{\bibinfo{booktitle}{Proceedings of AAAI 2019}}
  (\bibinfo{year}{2019}).

\bibitem{MEDIQA2019}
\bibinfo{author}{{Ben Abacha}, A.}, \bibinfo{author}{Shivade, C.} \&
  \bibinfo{author}{Demner{-}Fushman, D.}
\newblock \bibinfo{title}{Overview of the mediqa 2019 shared task on textual
  inference, question entailment and question answering}.
\newblock In \emph{\bibinfo{booktitle}{Proceedings of the 18th BioNLP Workshop and Shared Task}} 
\bibinfo{pages}{370–379} (\bibinfo{publisher}{Association for Computation Linguistics}, 
  \bibinfo{address}{Florence, Italy}
  \bibinfo{year}{2019}).

\bibitem{rouge}
\bibinfo{author}{Lin, C.-Y.}
\newblock \bibinfo{title}{{ROUGE}: a package for automatic evaluation of
  summaries}.
\newblock In \emph{\bibinfo{booktitle}{Text Summarization Branches Out}},
  \bibinfo{pages}{74--81} (\bibinfo{publisher}{Association for Computational
  Linguistics}, \bibinfo{address}{Barcelona, Spain}, \bibinfo{year}{2004}).

\bibitem{bleu}
\bibinfo{author}{Papineni, K.}, \bibinfo{author}{Roukos, S.},
  \bibinfo{author}{Ward, T.} \& \bibinfo{author}{Zhu, W.-J.}
\newblock \bibinfo{title}{BLEU: a method for automatic evaluation of machine
  translation}.
\newblock In \emph{\bibinfo{booktitle}{Proceedings of the 40th Annual Meeting
  of the Association for Computational Linguistics}}, \bibinfo{pages}{311–318}
  (\bibinfo{publisher}{Association for Computational Linguistics},
  \bibinfo{address}{USA}, \bibinfo{year}{2002}).

\bibitem{lstm}
\bibinfo{author}{Hochreiter, S.} \& \bibinfo{author}{Schmidhuber, J.}
\newblock \bibinfo{title}{Long short-term memory}.
\newblock \emph{\bibinfo{journal}{Neural computation}}
  \textbf{\bibinfo{volume}{9}}, \bibinfo{pages}{1735--80}
  (\bibinfo{year}{1997}).

\bibitem{chen2018fast}
\bibinfo{author}{Chen, Y.-C.} \& \bibinfo{author}{Bansal, M.}
\newblock \bibinfo{title}{Fast abstractive summarization with
  reinforce-selected sentence rewriting}.
\newblock In \emph{\bibinfo{booktitle}{Proceedings of the 56th Annual Meeting of the Association for Computational Linguistics}}
\bibinfo{pages}{675–686} (\bibinfo{publisher}{Association for Computation Linguistics}, 
  \bibinfo{address}{Melbourne, Australia}
  \bibinfo{year}{2019}).

\bibitem{Liu2019HierarchicalTF}
\bibinfo{author}{Liu, Y.} \& \bibinfo{author}{Lapata, M.}
\newblock \bibinfo{title}{Hierarchical transformers for multi-document
  summarization}.
\newblock In \emph{\bibinfo{booktitle}{Proceedings of the 57th Annual Meeting
  of the Association for Computational Linguistics}},
  \bibinfo{pages}{5070--5081} 
  (\bibinfo{publisher}{Association for
  Computational Linguistics}, 
  \bibinfo{address}{Florence, Italy},
  \bibinfo{year}{2019}).

\bibitem{lewis2019bart}
\bibinfo{author}{Lewis, M.} \emph{et~al.}
\newblock \bibinfo{title}{BART: denoising sequence-to-sequence pre-training for
  natural language generation, translation, and comprehension}.
\newblock \emph{\bibinfo{journal}{arXiv preprint arXiv:1910.13461}} (\bibinfo{year}{2019}).

\bibitem{devlin2018}
\bibinfo{author}{Devlin, J.}, \bibinfo{author}{Chang, M.-W.},
  \bibinfo{author}{Lee, K.} \& \bibinfo{author}{Toutanova, K.}
\newblock \bibinfo{title}{BERT: pre-training of deep bidirectional transformers
  for language understanding}.
\newblock In \emph{\bibinfo{booktitle}{Proceedings of the 2019 Conference of the North {A}merican Chapter of the Association for Computational Linguistics}},
  \bibinfo{pages}{4171--4186} (\bibinfo{publisher}{Association for Computation Linguistics}, 
  \bibinfo{address}{Minneapolis, Minnesota}
  \bibinfo{year}{2019}).

\bibitem{radford2019}
\bibinfo{author}{Radford, A.} \emph{et~al.}
\newblock \bibinfo{title}{Language models are unsupervised multitask learners}
\newblock \emph{\bibinfo{journal}{OpenAI Blog}} 
\newblock at https://openai.com/blog/better-language-models (\bibinfo{year}{2019}).

\bibitem{BenAbacha:MEDINFO19}
\bibinfo{author}{{Ben Abacha}, A.} \emph{et~al.}
\newblock \bibinfo{title}{Bridging the gap between consumers’ medication
  questions and trusted answers}.
\newblock In \emph{\bibinfo{booktitle}{MEDINFO 2019}} (\bibinfo{year}{2019}).

\bibitem{ctrl}
\bibinfo{author}{Keskar, N.~S.}, \bibinfo{author}{McCann, B.},
  \bibinfo{author}{Varshney, L.}, \bibinfo{author}{Xiong, C.} \&
  \bibinfo{author}{Socher, R.}
\newblock \bibinfo{title}{{CTRL - A Conditional Transformer Language Model for
  Controllable Generation}}.
\newblock \emph{\bibinfo{journal}{arXiv preprint arXiv:1909.05858}} (\bibinfo{year}{2019}).


\end{thebibliography}
%\bibliographystyle{naturemag}

\end{document}